\documentclass{article}

\usepackage{arxiv}

\usepackage[utf8]{inputenc} 
\usepackage[T1]{fontenc}    
\usepackage[greek, english]{babel}
\usepackage{hyperref}       
\usepackage{url}            
\usepackage{booktabs}       
\usepackage{amsfonts}       
\usepackage{nicefrac}       
\usepackage{microtype}      
\usepackage{thmtools}
\usepackage{float}
\usepackage[round, sort]{natbib}
\usepackage{graphicx}
\graphicspath{ {./images/} }

\declaretheorem{Definition}

\title{Using Finite-State Machines to Automatically Scan Classical Greek Hexameter}

\author{
 Anne-Kathrin Schumann \\
  T2K: Text to Knowledge\\
  Dresden, Germany\\
  \texttt{anne-kathrin.schumann@text2knowledge.de} \\
   \And
 Christoph Beierle \\
  Faculty of Mathematics and Computer Science, Knowledge-Based Systems\\
  FernUniversität in Hagen\\
  Hagen, Germany\\
  \texttt{christoph.beierle@fernuni-hagen.de} \\
  \And
 Norbert Blö{\ss}ner \\
  Institute of Greek and Latin Languages and Literatures\\
  Freie Universität Berlin\\
  Berlin, Germany\\
  \texttt{n.bloessner@fu-berlin.de} \\
}

\begin{document}
\maketitle
\begin{abstract}
This paper presents a fully automatic approach to the scansion of Classical Greek hexameter verse. In particular, the paper describes an algorithm that uses deterministic finite-state automata and local linguistic rules to implement a targeted search for valid spondeus patterns and, in addition, a weighted finite-state transducer to correct and complete partial analyses and to reject invalid candidates. The paper also details the results of an empirical evaluation of the annotation quality resulting from this approach on hand-annotated data. It is shown that a finite-state approach provides quick and linguistically sound analyses of hexameter verses as well as an efficient formalisation of linguistic knowledge. The project code is available online\footnote{\url{https://github.com/anetschka/greek\_scansion}.}. 
\end{abstract}


\section{Introduction}
Greek literature has, for centuries, served as a paradigm and model for literary writing all over Europe. The oldest surviving texts of Classical Greek literature -- texts such as the Iliad, the Odyssey, and the works of Hesiod -- are epic poems that all share the same metre: hexameter. They are written in an artificial language that has never been spoken in everyday life and owes its origin and many of its peculiarities to the nature of metrically bound language (\cite{meister1921}). Comprehensive hexameter annotation is, therefore, crucial for large-scale and data-driven investigations into some of the linguistic features of Ancient Greek epic language.

Furthermore, it may provide additional criteria for the evaluation of Homer's repeated verses, the so-called \textit{iterata}. Within Classical Philology, controversy around the nature of the Homeric repetitions started in 1840, and it remained one of the central research questions in the field for a long period of time (see \citet{strasser}, pp. 1-5). However, after Milman Parry (\citeyear{parry}) had shown that a certain amount of the \textit{iterata} may be well explained as \textit{formulae} -- repetitions from an oral corpus that served as starting points for improvisations by the oral singer (the \textit{Aoidos}) --, scientific interest in alternative views on the \textit{iterata} declined (see \citet{west}, pp. 3-5), particularly among English-speaking scholars. Until today, numerous specialists adhere to the view that all \textit{iterata} are also \textit{formulae} (see, for instance, \cite{pavese}). Empirical analyses, however, do not support this view (\citet{strasser}, pp. 37-40). 

Large-scale hexameter scansion may help to distinguish between \textit{formulae} and \textit{non-formulae} and, thus, foster new, empirically well-founded views on one of the core research questions in Classical Philology, i.e., the ``Homeric question'' (see \cite{blosner1} for the current state of research on this topic).

\section{Classical Greek Hexameter}\label{hexameter}

\subsection{Quantitative Metre}

Unlike metric systems of modern European languages, Greek hexameter is built not on stress patterns, but exploits the properties of the Ancient Greek vowel system: Ancient Greek was a tonal language that distinguished between long and short vowels or diphthongs, respectively. Consequently, the Ancient Greek metric system is \textit{quantitative}, i.e., at the basis of the metre is not a stress sequence, but a sequence of long and short syllables, and syllable duration is determined by the syllable’s vowel or diphthong. 

It follows that Greek hexameter verses can be described as regular sequences of long and short syllables. In hexameter, long and short syllables are organised in feet of the following form: 

\begin{itemize}
    \item \textit{Dactyl}: The foot is composed of three syllables, the first of which is long, while the others are short (pattern \textit{long-short-short}). 
    \item \textit{Spondee}: The foot is composed of two long syllables (pattern \textit{long-long}).
    \item The first syllable of each foot is always long, and six feet make a complete hexameter.
\end{itemize}

\begin{figure}
    \centering
    \includegraphics[width=0.8\textwidth]{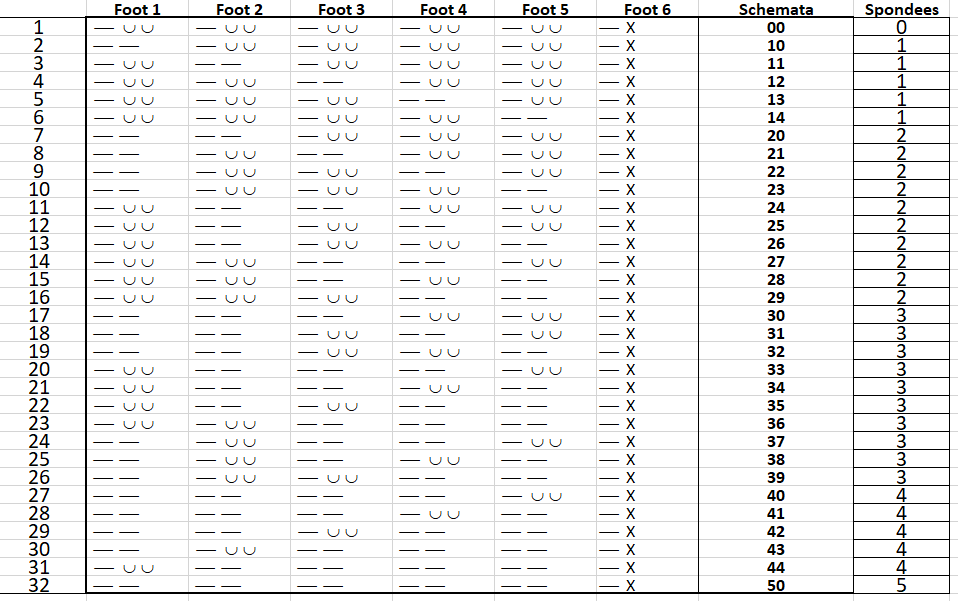}
    \caption{The 32 variants of Ancient Greek hexameter (\cite{hoflmeier1982}). Horizontal bars indicate long syllables, bows indicate short syllables. X marks the \textit{anceps}. The variants are indexed according to the number of spondees in the verse. For instance, the index 50 indicates that the verse has five spondees.}
    \label{hexameterschema}
\end{figure}

The difficulty of hexameter scansion consists in determining the distribution of dactyls and spondees over feet. This distribution is controlled by both local and global parameters. We call ``local'' those parameters that describe the immediate linguistic context of a given syllable, whereas ``global'' are those parameters that refer to the overall structure of the hexameter verse. The following global parameters are important for hexameter scansion: 

\begin{itemize}
    \item In a valid hexameter verse, feet one to five are allowed to be either spondees or dactyls\footnote{However, philologists know by experience that the probability for a given foot to be a dactyl is not uniformly distributed.}, only the last foot is restricted with respect to its metric form: It is composed of a long syllable and a second one which can be either long or short (the so-called \textit{anceps}).
    \item How many spondees and dactyls a hexameter verse has, depends on the syllable count. The minimal length of a hexameter is twelve syllables: five spondees (i.e. two-syllable feet) plus the two syllables of the last foot (see hexameter no. 50 in Figure \ref{hexameterschema}). The maximum length of a hexameter verse is 17 syllables: five dactyls (i.e. three-syllable verses) plus the two syllables of the last foot (see hexameter no. 00 in Figure \ref{hexameterschema}). Figure \ref{hexameterschema} is a graphical depiction of the resulting 32 variants of Classical Greek hexameter (following \citet{hoflmeier1982}).
\end{itemize}

\subsection{Length of Vowels and Diphthongs}

Moreover, the distribution of spondees and dactyls over the feet of a hexameter verse must comply with local parameters, i.e. the prosodic values of the verse’s syllables. These are traditionally modelled by means of linguistic rules that are local in the sense that they describe only the immediate, local context of vowels and diphthongs. In our work, we use the following generalisation of traditional philological rulesets (see also \citet{leggewie}, \citet{papakitsos}): 

\begin{enumerate}
    \item \textit{Long by nature (1)}:  The vowels eta \textgreek({H}, \textgreek{η}) and omega (\textgreek{Ω}, \textgreek{ω}) are long by nature.
    \item \textit{Long by nature (2)}: Most valid diphthongs are long by nature. \citet{papakitsos} provides more fine-grained information.
    \item \textit{Long by position}: Vowels followed by pairs of consonants -- or one of the phonologically ``complex'' consonants zeta (\textgreek{ζ}), xi (\textgreek{ξ}), and psi (\textgreek{ψ}) -- are long by position.
\end{enumerate}

Exceptions from these rules can occur in the following cases: 

\begin{itemize}
    \item \textit{Sequences of vowels}: Sequences of two vowels, or of a vowel and a diphthong, can cause exceptions from the above-mentioned rules. An example is \textit{hiatus} which occurs when the last vowel of a word and the initial vowel of the next word are directly adjoined. In most cases, \textit{hiatus} causes the first vowel to be substituted by an apostrophe (\textit{elision}), if it is short. If the vowel is long, \textit{hiatus} can lead to its prosodic shortening. Within a word, sequences of vowels can be merged into one (mostly long) vowel, a phenomenon that is called \textit{synizesis}. A well-known example is the first verse of the Iliad, given in Listing 1. Here, the vowels \textgreek{ε} and \textgreek{ω} are fused (together with the preceding consonant) into one long syllable \textgreek{δεω}. \textit{Crasis} is a similar phenomen that, however, crosses word boundaries. 
    
    1) IL.1,1 \textgreek{Μ\~{η}νιν \accpsili\'{α}ειδε θε\`{α} Πηλη\"{ι}\'{α}\underline{δεω} \accpsili{}Αχιλ\~{η}ος}
    
    \item \textit{Muta cum liquida}: These are sequences of ``mute'' and ``liquid'' consonants. ''Liquid'' consonants are lambda (\textgreek{λ}), rho (\textgreek{ρ}), my (\textgreek{μ}), and ny (\textgreek{ν}). ``Mute'' consonants can be tenuis, voiced, or aspirated. \textit{Muta cum liquida} can cause exceptions from positional lengthening. 
\end{itemize}

\subsection{Diacritics}

Ancient Greek texts have, in later times, been marked up with diacritics to make reading easier. Some of these diacritics provide information that may help hexameter scansion:  

\begin{itemize}
    \item The \textit{trema} signals that a vowel is not fused with a preceding vowel into a diphthong, but forms a syllable of its own. An example is the formula \textgreek{Πηλη\"{ι}\'{α}δεω \accpsili{}Αχιλ\~{η}ος} (``Achilles, son of Peleus'') in Listing 1, where the \textit{trema} on vowel iota (\textgreek{\"{ι}}) signals that the vowel is not fused with its predecessor into the diphthong \textgreek{ηι}. 
    \item The \textit{circumflex} marks a vowel as being long. An example can, again, be found in the formula \textgreek{Πηλη\"{ι}\'{α}δεω \accpsili{}Αχιλ\~{η}ος}, where the \textit{circumflex} on vowel \textgreek{\~{η}} indicates prosodic length.
\end{itemize}

\subsection{Summary}

Automatic Greek hexameter scansion is the task of determining: 

\begin{enumerate}
    \item whether or not a given verse is a valid hexameter verse, and
    \item of finding a sequence of spondees and dactyls that complies both with global and local wellformedness rules.
\end{enumerate}

Additional information can be drawn from diacritics and the fact that spondees and dactyls are not evenly distributed over feet. For instance, \citet{papakitsos} points out that the fifth foot of a verse, by experience, has the highest probability of being a dactyl, whereas the second foot has the lowest probability of being a dactyl. 

\section{Related Work}\label{relatedwork}

An early, rule-based approach to the semi-automatic scansion of Greek hexameter has been developed by \citet{hoflmeier1982}. Höflmeier combines two different kinds of knowledge to resolve hexameter verses: 

\begin{itemize}
    \item Local linguistic rules that establish which vowels are short, and which vowels are long.
    \item Knowledge about the overall structure of the verse for the resolution of partially annotated verses. 
\end{itemize}

Höflmeier developes a 2000-line Pascal program that implements the scansion algorithm. The program scans verses left-to-right and explicitly encodes linguistic rules in the form of Pascal procedures. The approach is semi-automatic since the resolution of synizesis and irregular cases is delegated to the user. A similar approach has later been proposed by \citet{pavese} who use series of regular expression substitutions to semi-automatically scan Greek verses. The authors report that this methodology allowed them to scan about 90 \% of all verses in their corpus. However, they do not provide an evaluation of the precision of their scansion.

A technologically more advanced study in the automatic scansion of metric poetry is the work by \citet{greene} who use weighted finite-state transducers, trained on a small corpus of manually annotated data, to analyse Shakespearean sonnets. The authors report accuracy values of up to 81.4 \%, achieved on a very small test set of 70 verses. 

An interesting approach to the problem of Greek hexameter scansion is presented by \citet{papakitsos}. Papakitsos performs syllabification and then employs a search strategy to identify dactyls, that is, the verses are not analysed left-to-right. Rather, the search starts in the fifth foot where dactyls are particularly likely, and dactyls are identified by means of local linguistic rules.  Once the appropriate – for the established number of syllables in the verse – number of dactyls has been identified, the search terminates. The search, however, is strongly dependent on the correctness of the syllabification. For instance, if the verse under analysis has been found to consist of 13 syllables (i.e., it corresponds to one of the variants given under indexes 40-44 in Fig. \ref{hexameterschema}), the search algorithm will look for exactly one dactyl. Papakitsos reports a recall of 0.98 and a precision of 0.80, evaluating on 700 verses. 

Linguistically informed approaches to the scansion of metric poetry have also been proposed by \citet[for Spanish sonnets]{sonnets} and \citet[for English]{hulden}. However, these studies heavily rely on lexical stress and are, therefore, not relevant for Greek hexameter. 

A rule-based implementation of a fully automatic Greek hexameter scansion algorithm has been published by Hope Ranker\footnote{\url{https://github.com/epilanthanomai/hexameter}.}. This algorithm uses an ensemble of weighted finite-state transducers to resolve the feet one by one. 

Due to the scarcity of data and the regular nature of metric poetry, machine learning has rarely been applied to the task of poetry scansion. When it is employed, the problem is usually modelled as syllable-wise classification. For instance, \citet{hench3} use a Conditional Random Fields classifier to analyse Middle High German epic texts, reaching an f-measure of 0.90 for syllable-wise evaluation, on 75 verses. \citet{zabaletak2017} reports on a very wide range of experiments, but achieves the best results (95 \% accuracy, evaluated syllable-wise on 78 poems) with a combination of a sequential model and deep learning for the classification of English, Spanish, and Basque verses. N-grams, positional and length features as well as linguistic markers are used to train the models. However, the very small data sets used for evaluation in these studies leave room for doubts with respect to the generalisation abilities of the proposed models. 

\section{Finite-State Approach to Hexameter Analysis}

\subsection{Motivation}

Our approach to the scansion of Classical Greek hexameter is fully automatic and rule-based. It uses linguistic rules and finite-state technology to verify the validity of hexameter candidates and to assign a scansion (as per Fig. \ref{hexameterschema}) to each valid candidate. The decision to explicitly encode linguistic knowledge is motivated not only by the internal requirement to provide a linguistically transparent solution (i.e., rule application), but also by the lack of annotated training data. Moreover, some of the linguistic properties of Greek epic writing do not favour the use of machine learning, particularly in a setting where no linguistic resources are available: Ancient Greek has a rich morphology. Furthermore, it is a well-established fact (\cite{hackstein, hackstein2, homer}) that the texts under study do not represent a ``natural'', but a rather artificial language in which linguistic innovation and archaic linguistic structures co-occur side by side. Given the so far inconsistent performance of machine learning approaches on poetry scansion (see Section \ref{relatedwork}), there is reason to believe that leveraging the expert knowledge that philologists have accumulated over centuries of research may be beneficial for automatic scansion.

In fact, the use of finite-state technology for the resolution of Classical Greek hexameter seems quite natural. Linguistic work on Greek hexameter has traditionally employed finite sets of local and global linguistic rules (see Section \ref{hexameter}). Embedding these two kinds of knowledge into finite-state automata seems relatively straightforward. Furthermore, finite-state technologies can be used both in low-resources scenarios and in situations in which annotated corpora are available (i.e., for training finite-state transducers). An additional advantage of finite-state automata is that they can be visualised in the form of state diagrams and thus provide a means for analysing rule application – and the validity of rulesets – in the philology classroom as well as in philological research.

\subsection{Definitions}

\begin{figure}
    \centering
    \includegraphics[width=0.6\textwidth]{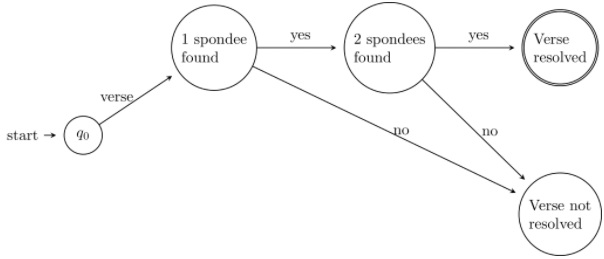}
    \caption{Example of a simple deterministic finite-state automaton.}
    \label{deafig}
\end{figure}

We use two types of finite-state machines, namely deterministic finite-state automata as well as a weighted finite-state transducer.

\begin{Definition}[Deterministic finite-state automaton]\label{dea}
A deterministic finite-state automaton is a tuple $(Q, \Sigma, \delta, q_{0}, F)$, such that: 
\begin{itemize}
	\item $Q$ is a finite, non-empty set of states, 
	\item $\Sigma$ is a finite alphabet,
	\item $\delta : Q \times \Sigma \rightarrow Q$ is the transition function,
	\item $q_{0} \in Q$ is the initial state,
	\item $F \subseteq Q$ is the set of accepting states.
\end{itemize}
\end{Definition}

Deterministic finite-state automata process inputs by iterating over each individual element of a given input. The transition function deterministically, i.e. univocally, defines which state the automaton will reach after processing an input element in a given state. The automaton is said to ``accept'' an input word if the process terminates in an accepting state $q \in F$. In computing, finite-state automata can be used to enforce wellformedness conditions.

Fig. \ref{deafig} is an example of a deterministic finite-state automaton. There are five states. The Greek alphabet (plus the words ``yes'' and ``no'') constitutes the input alphabet. $\delta$ is given by the transitions between states, given a specific input. The state ``Verse resolved'' is the accepting state. This automaton represents a na\"{i}ve implementation of a scansion automaton that accepts verses with two spondees, i.e., it can decide whether input verses of 15 syllables (indexes 20-29 in Fig. \ref{hexameterschema}) are valid hexameter candidates. The representation is simplified since the state-internal routines that are responsible for finding spondees in given positions are omitted\footnote{A more realistic automaton could, for instance, use sub-states to search for spondees in all relevant positions.}. To output a scansion, the automaton would need to take note of the positions of the spondees and fill the remaining positions with dactyls. 

\begin{figure}
    \centering
    \includegraphics[width=0.4\textwidth]{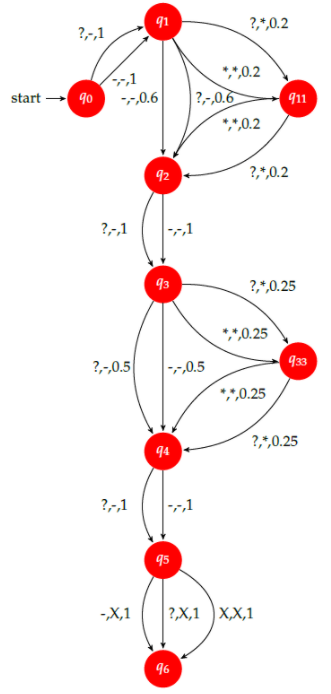}
    \caption{Example of a weighted finite-state transducer.}
    \label{fstfig}
\end{figure}

\newpage

\begin{Definition}[Finite-State Transducer]\label{fst}
A finite-state transducer is a tuple $(\Sigma_{1}, \Sigma_{2}, Q, q_{0}, F, E)$, such that:
\begin{itemize}
    \item $Q$ is a finite, non-empty set of states,
    \item $\Sigma_{1}$ is a finite input alphabet,
    \item $\Sigma_{2}$ is a finite output alphabet,
    \item $q_{0} \in Q$ is the initial state,
    \item $F \in Q$ is the set of accepting states,
    \item $E$ is the set of edges.
\end{itemize}
\end{Definition}

Unlike deterministic finite-state automata, finite-state transducers output an element $o \in \Sigma_{2}$ for each input element $i \in \Sigma_{1}$. By doing so, a finite state transducer creates a relation between words $w_{1} \in \Sigma_{1}$ and $w_{2} \in \Sigma_{2}$. For instance, a finite-state transducer can theoretically be used to translate a given hexameter verse into one of the hexameter schemes in Fig. \ref{hexameterschema}. Another important characteristic of finite-state transducers is that they are not necessarily deterministic. This means that there may be several valid paths of an input word through the finite-state transducer. For instance, a finite-state transducer that translates Classical Greek hexameter verses into hexameter schemes as per Fig. \ref{hexameterschema} might, for a given verse, calculate several plausible variants. In this situation, the addition of weights to the edges of the transducer allows for the ranking of variants. 

Fig. \ref{fstfig} shows an example of a weighted finite-state transducer. Each edge is labelled with an input symbol $i \in  \Sigma_{1}$, an output symbol $o \in \Sigma_{2}$, and an edge weight. The figure also shows that there are several paths that end in the accepting state $q_{6}$, that is, the transducer is non-deterministic. 

\subsection{Algorithm}

\begin{figure}
    \centering
    \includegraphics[width=0.6\textwidth]{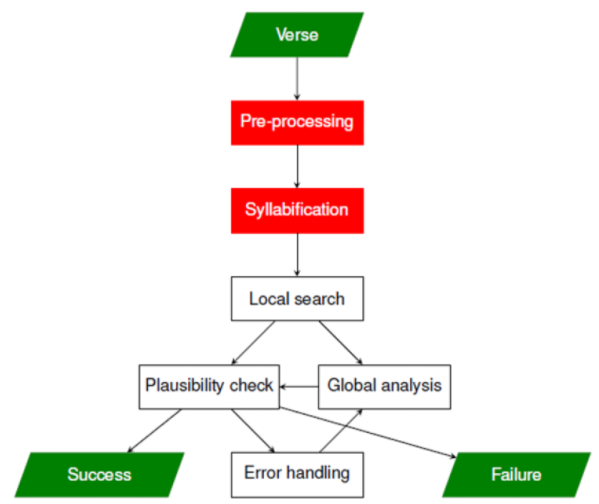}
    \caption{Visual representation of the scansion algorithm.}
    \label{scansionfig}
\end{figure}

Our approach to the scansion of Ancient Greek hexameter is based on the same two types of knowledge that were already used by \citet{hoflmeier1982}: 

\begin{itemize}
    \item \textit{Local search}: We use the local linguistic rules given in Section \ref{hexameter} to determine whether a pair of syllables can safely be considered long (that is, it forms a spondeus). This is implemented with the help of deterministic finite-state machines. 
    \item \textit{Global analysis}: We exploit knowledge about the overall structure of Greek hexameter to complete partially annotated verses, that is, verses that could not fully be resolved with the help of the linguistic rules. This is done using a finite-state transducer. 
\end{itemize}

The local search step follows the strategy of \citet{papakitsos} in that it searches for a fixed number of spondees that result from the syllable count established during syllabification. Fig. \ref{scansionfig} shows a visual representation of our scansion algorithm. The algorithm scans epic Greek text verse by verse: 

\begin{enumerate}
    \item \textit{Pre-processing} consists mainly of lower-casing and the removal of (uninformative) diacritics. 
    \item Moreover, we have implemented a \textit{syllabification} algorithm that uses regular expressions to identify syllables and to establish the syllable count of the verse.
    \item The \textit{local search} and all following steps are then handled by dedicated deterministic finite-state automata. There are specialised automata for verses of 13, 14, 15, and 16 syllables and a simpler FSA for all remaining cases. In the local search step, the active automaton performs a targeted search for a given number of spondees, using the local linguistic rules. If enough spondees are found, the plausibility of the solution is checked. Otherwise, the verse is passed to the global analysis step. The automata were implemented using an existing Python library\footnote{\url{https://github.com/pytransitions/transitions}.}. 
    \item For \textit{global analysis}, we use a non-deterministic finite-state transducer. In this transducer, each syllable corresponds to a state, and alternative plausible solutions are modelled by means of alternative paths. The transducer is weighted, but since we did not have access to an appropriate training corpus, we were not able to learn transition weights from data. Instead, they were set manually following the description provided by \citet{papakitsos}. The transducer was implemented using the Helsinki Finite-State Tools\footnote{\url{ https://hfst.github.io/python/3.12.1/index.html}.}. 
    \item If the \textit{plausibility check} or the \textit{global analysis} fail, the verse is passed over to \textit{error handling} to compensate for potentially erroneous syllabification. Global analysis then completes the verse. The plausibility of the result is checked again. Depending on this result, the FSA will transition to its final state, that is, either \textit{success} or \textit{failure}.  
\end{enumerate}

If the verse, however, passes the plausibility check immediately after the local search step, the finite-state automaton transitions directly to the \textit{success} state. 

\subsection{Local Search and Linguistic Rules}

\begin{figure}
    \centering
    \includegraphics[width=0.4\textwidth]{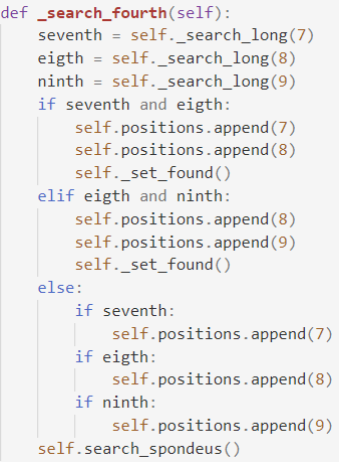}
    \caption{Python code of the procedure \textit{search\_fourth} of the deterministic finite-state automaton that handles verses of 13 syllables.}
    \label{searchfourth}
\end{figure}

According to the annotation algorithm in Fig. \ref{scansionfig}, each verse is handed over to a dedicated finite-state automaton after syllabification. This automaton handles the targeted search for the correct number of spondees. For instance, the automaton that handles 16-syllable verses (indexes 10-14 in Fig. \ref{hexameterschema}) will search for exactly one spondee. Following \citet{papakitsos}, the search is targeted in the sense that it is not carried out from left to right, but starts in the second foot where spondees are known to be particularly likely. Since no foot boundaries are known at this point, all syllables that are potential members of a given foot are considered in the search.  

An example of how this is done is shown in Fig. \ref{searchfourth}. The figure shows the procedure that searches for a spondee in the fourth foot of verses with 13 syllables. Syllables 7-9 are considered for this search. Linguistic rules as per Section \ref{hexameter} are applied to determine, for each syllable, whether it can safely be considered long (call to the function \textit{search\_long}). If two consecutive long syllables are identified, the automaton decides that a spondee is found (call to the procedure \textit{set\_found}) and the automaton will proceed to search for the next spondeus in another foot. If, however, no such pair is found, the automaton takes note of the identified lengths\footnote{This information will later be used as a partial annotation in global analysis.} and then continues to search for the same spondeus in another foot. Once the required number of spondees is found, the search stops.  

A set of only five linguistic rules is used to search for syllable pairs that can safely be considered spondees. These rules are implemented using regular expressions. All rules are logically concatenated as shown in predicate logic in equation~1: 

\begin{equation}\label{finalequation}
R_{zf}(X) \lor R_{nl1}(X) \lor R_{nl2}(X) \lor (R_{pl}(X) \land \neg R_{ml}(X)) \Rightarrow isLong(X)
\end{equation}

Each predicate in the formula in equation 1 directly refers to one of our rule implementations. Precisely, $R_{zf}$ implements length based on the circumflex marker. $R_{nl1}$ and $R_{nl2}$ implement natural length, whereas $R_{pl}$ and $R_{ml}$ implement positional length and \textit{muta cum liquida}, respectively. 

\subsection{Global Analysis}

In the global analysis step, knowledge about the global structure of hexameter is used to complement the local linguistic rules when rule application alone is not sufficient to resolve a verse. In particular, a finite-state transducer is used to decide whether or not the input verse is a valid hexameter at all, and to complete partial analyses. As has been shown in the previous section, rule application can result in partial analyses if the finite-state automaton has found individual lengths, but not enough pairs of long syllables (i.e., spondees).  

A shortened representation of the finite-state transducer is shown in Fig. \ref{fstfig}. In this transducer, each syllable corresponds to a state. Alternative (plausible) solutions and varying syllable counts are modelled by means of alternative paths through the transducer. The state $q_{6}$ is the accepting state, and if no path through the transducer can be found for a given verse, the verse will be considered invalid (for instance, it could be a fragment). Transducer weights could not be learned, but were set following the explanation found in \citet{papakitsos}.  

As can be seen in the figure, processing starts in state $q_{0}$. States $q_{1}$, $q_{2}$, and $q_{11}$ represent the first foot of the verse. States $q_{5}$ and $q_{6}$ process the last foot of the verse. The body of the verse is processed by the remaining states\footnote{The representation of the transducer is shortened due to lack of space. In the real transducer, states $q_{3}$, $q_{4}$, and $q_{33}$ are repeated four times (for feet 2, 3, 4, and 5), but each time with different weights.}. The transducer replaces scansion symbols resulting from local search or error handling to come up with a plausible and complete scansion of the verse. In these replacements, $?$ signifies that the length of the syllable is unknown, whereas – and $*$ mark long and short syllables, respectively. If at least one solution is available for a given verse, i.e. if the input verse is valid, all question marks will be replaced by valid scansion symbols at the end of the analysis. 

\subsection{Error Handling}

Error handling accounts for the possibility of incorrect syllabification and the consequent incorrect application of the finite-state automata. It also includes a very simple, but incomplete, routine for identifying \textit{synizesis}. The error handling step does not use the syllabification result, but tries to establish for each vowel individually whether it is long or short.  The result of this process is then completed and corrected by the finite-state transducer. If, however, the transducer fails to output a valid result (i.e., if it rejects the input string), the error correction step tries to identify a possible \textit{synizesis} as shown in Listing 1. 

\section{Experiments}

\subsection{Corpus of Classical Greek Poetry}

\begin{table}
	\centering\scalebox{0.9}{
		\begin{tabular}{l|l}
			\textbf{Subcorpus} & \textbf{Number of verses}\\\hline
			Iliad & 15.875\\
			Odyssey & 12.150\\
			Hesiod, Theogony & 1066\\
			Hesiod, Erga & 840\\
			Pseudo-Hesiod, Aspis & 487\\
			Pseudo-Hesiod, Fragments & 1835\\
			Homeric Hymns & 2344\\
			Oedipodeia, Oechalias & 3\\
			Epigonoi & 1\\
			Cypria & 51\\
			Aethiopis & 2\\
			Ilias parva & 48\\
			Nostoi & 9\\
			Varia & 39\\
			\hline
			 & \textbf{$\Sigma$ 34.750}\\
		\end{tabular}}
		\caption{Corpus of Classical Greek poetry.}
		\label{vert}
\end{table}

For the development and evaluation of the scansion algorithm, we used a corpus of Classical Greek poetry that was initially collected at the University of Regensburg in the 1980s (see \citet{hoflmeier1982, strasser, blosner}). The corpus contains not only the verses themselves, but also a complete list of all iterata. For our work, we exported the verses from a MySQL database that was created by \citet{kruse}. Table \ref{vert} gives an overview of this dataset. 

\subsection{Results}

We have evaluated the performance of both our syllabification and our scansion module against hand-annotated verse data. The annotations were carried out by two advanced students of Greek philology both for syllabification and the actual hexameter scansion. Discrepancies and errors were clarified by means of group discussions. For syllabification evaluation, we randomly chose a set of 171 verses (2695 syllables) from both the Odyssey and the Iliad. For scansion evaluation, we randomly selected 346 verses from a broader range of Greek texts. Table \ref{datentab} provides an overview of the evaluation data set. 

Scansion correctness was evaluated by means of precision, recall, and f-measure as shown in the following definitions:

\begin{Definition}[Precision]\label{prec}	
	$p = \frac{t_{p}}{t_{p} + f_{p}}$
\end{Definition}

\begin{Definition}[Recall]\label{rec}
	$r = \frac{t_{p}}{t_{p} + f_{n}}$
\end{Definition}

\begin{Definition}[F-Measure]
	$F = \frac{2*pr}{p + r}$
\end{Definition}

Precision measures the correctness of all annotations provided by the algorithm: Correct annotations (``true positives'', $t_{p}$) are divided by the sum of correct and incorrect annotations (“false positives”, $f_{p}$). Since perfect precision can theoretically be achieved on a very small data set and is, therefore, insufficient for estimating the true quality of a scansion algorithm, recall relates correct annotations to the amount of those verses for which the algorithm did not output any solution (``false negatives'', $f_{n}$). The f-measure then combines both scores into a unified performance score that can be compared across different systems. The evaluation scripts are included in the open-source code package of our software. 

\begin{table}[b]
	\centering
	\begin{tabular}{l|c|c|c|c|c|c|c|c}
		 & \textbf{Iliad} & \textbf{Odyssey} & \textbf{Theogony} & \textbf{Erga} & \textbf{Aspis} & \textbf{Fragments} & \textbf{Hymns} & \textbf{Total}\\\hline
		Syllabification & 159 & 12 & -- & -- & -- & -- & -- & \textbf{171}\\
		Scansion & 159 & 122 & 11 & 8 & 5 & 18 & 23 & \textbf{346}\\
	\end{tabular}
	\caption{Data set used for evaluation.}
	\label{datentab}
\end{table}

\subsubsection{Syllabification}

Since correct syllabification is an important prerequisite for our algorithm to work properly, we evaluated the correctness of this step using accuracy, i.e., the percentage of correct syllable splits over our test set of 171 verses (see Table \ref{datentab}). The 171 verses tested contain 2695 syllables. We achieved a verse-wise accuracy of 0.82 and a syllable-wise accuracy of 0.98.

The relatively large margin between syllable-wise and verse-wise evaluation indicates that many cases of incorrect syllabification are due to single, and local, errors. We carried out an error analysis and concluded that the larger part of incorrect splits is caused by mistakes in the assignment of individual consonants to the sequence either before or after the split. However, this kind of mistake does not affect the resulting syllable count. For better verse-wise accuracy, it would have been necessary to encode higher levels of linguistic knowledge or to hand-annotate a sufficient amount of data for the training of a classification algorithm. For our low-resource setting, we, however, deemed our syllabification approach to be ``good enough'' for the scansion task.

\subsubsection{Hexameter Scansion}

\begin{table}
	\centering
	\begin{tabular}{l|c|c|c}
		 & \textbf{Precision} & \textbf{Recall} & \textbf{F-Measure}\\\hline
		Local search only & 0.97 & 0.66 & \textbf{0.78}\\
		Local search + global analysis & 0.94 & 0.97 & \textbf{0.96}\\
		Baseline & 0.98 & 0.98 & \textbf{0.98}\\
		Complete algorithm & 0.95 & 1.00 & \textbf{0.98} \\
	\end{tabular}
	\caption{Results of scansion evaluation.}
	\label{evalresults}
\end{table}

We have evaluated the quality of our scansion algorithm against the open-source implementation provided by Hope Ranker on GitHub\footnote{\url{ https://github.com/epilanthanomai/hexameter}.}. Table \ref{evalresults} details the results for various forms of our algorithm: 

\begin{itemize}
    \item Local search with linguistic rules only, 
    \item Local search + global analysis, using the finite-state transducer, 
    \item The complete algorithm with local search, global analysis, and error handling.
\end{itemize}

The table proves the competitiveness of our approach with the baseline. Both the baseline and the full version of our algorithm outperform the approaches mentioned in Section \ref{relatedwork}. The table also shows that rule application alone results in very high precision, but insufficiently low recall. The addition of the finite-state transducer massively improves the performance of the algorithm and falsifies the claim made by \citet{hoflmeier1982} that only 90 \% of the verses can be automatically analysed.  

\subsubsection{Scansion of the Complete Corpus}

In another evaluation step, we have applied our finite-state algorithm to the complete corpus (see Table \ref{vert}) and compared the output to the output of Höflmeier’s (1982) Pascal program\footnote{It should be noted that Höflmeier’s approach is not fully automatic and leaves the resolution of problematic verses to the user. For our comparison, missing solutions were provided by students of Classical Philology.}. We then used the $\kappa$-coefficient (\citet{cohen}, cf. \citet{agreement}) to quantify the agreement between both ahttps://www.overleaf.com/project/5fdca08fdff08800ffb7241cnnotations as shown in equation \ref{kappa}: 

\begin{Definition}[Kappa]\label{kappa}
	$\kappa = \frac{A_{o} - A_{e}}{1 - A_{e}}$
\end{Definition}

In equation \ref{kappa}, $A_{o}$ refers to the agreement that was empirically observed between annotators. This measure is then compared to the expected agreement $A_{e}$, a statistical measure that is valid under the condition of completely random annotations. $\kappa$, therefore, measures the agreement that goes beyond random overlap. $\kappa$ values above 0.8 are traditionally considered as indicators of high, systematic agreement between annotators and, consequently, good annotation quality. We also evaluated the output of our algorithm by means of precision, recall, and f-measure, considering Höflmeier’s result as the gold standard annotation. The following results were obtained: 

\begin{itemize}
    \item Analysis of 34,599 verses
    \item Precision: 0.89
    \item Recall 1.0
    \item F-measure: 0.94 
    \item $\kappa$: 0.88 
\end{itemize}

The agreement between the Höflmeier program and our algorithm is high. An analysis of the disagreements shows that in cases, in which the verse under analysis is not a valid hexameter (but, for instance, a fragment), the Höflmeier algorithm outputs a stub, whereas our algorithm, correctly, outputs no scansion\footnote{In fact, precision rises to 0.92 when these cases of false hexameter candidates are disregarded.}. The analysis indicates that those annotations for which there is an agreement between both algorithms can, with some caution, be considered of high quality. 

\section{Conclusion}

In this paper, we have presented a fully automatic approach to the analysis of Ancient Greek hexameter. Automatic annotation tools are crucial for data-driven investigations in Greek philology. Our algorithm integrates various kinds of linguistic knowledge into a set of finite-state automata and thus makes use of well-defined concepts in the field of computational linguistics, while remaining transparent to philologists. Our evaluation results are competitive.  

One of the main advantages in the present work is the computationally efficient formalisation of linguistic knowledge: 

\begin{itemize}
    \item The scansion algorithm has been designed following best practices in software engineering, and it has been empirically evaluated. The code is open source. This opens up new opportunities for validated improvement of both the algorithm and the annotation quality. Both aspects are crucial for progress in empirical research in Greek philology. 
    \item We have shown that finite-state technology is helpful in the resolution of Classical Greek hexameter. Finite-state automata formalise linguistic knowledge transparently and in a way that can be understood by linguists. State diagrams of finite-state machines can be used in didactic settings and to verify annotation decisions. 
\end{itemize}

We have implemented a very small, and very simple, linguistic rule set: Neither was there a need for modelling complex interactions between rules, nor was it necessary for hexameter resolution to incorporate knowledge on more complex linguistic levels (e.g., semantics). However, hexameter scansion is not yet resolved: \textit{Synizesis} or \textit{crasis} cannot yet be discretely modelled. As soon as large-scale annotated hexameter corpora become available, more complex approaches to hexameter analysis should be explored. Statistical and machine learning approaches can provide new insights into the language of the Greek epos, and genetic algorithms could be used to learn optimal rule sets and, thus, to validate traditional philological approaches to hexameter. 

\section{Acknowledgement}

The authors would like to thank Jonas S\"{u}ltmann and Simon Neumaier for hand-annotating a large amount of development and test data. 

\bibliographystyle{abbrvnat}  
\bibliography{references}  

\end{document}